\documentclass{article}
\usepackage{spconf,amsmath,graphicx}

\usepackage{booktabs}
\usepackage{fixmetodonotes}
\usepackage[table, dvipsnames]{xcolor}
\usepackage{hyperref}
\usepackage{cleveref}
\usepackage{amsfonts}
\usepackage{enumitem}
\usepackage{multirow}
\usepackage{cancel}
\usepackage{makecell}
\usepackage{tikz}
\usetikzlibrary{shapes, arrows, positioning, fit, calc}

\crefname{figure}{Fig.}{Figs.}
\Crefname{figure}{Figure}{Figures}
\crefname{section}{Sec.}{Secs.}
\Crefname{section}{Section}{Sections}
\crefname{table}{Tab.}{Tabs.}
\Crefname{table}{Table}{Tables}

\usepackage{ifthen}

\newcommand{\condensedparagraph}[1]{\noindent\textbf{#1}\quad{}}

\usepackage{pifont}
\definecolor{MyGreen}{RGB}{0, 180, 0}
\definecolor{MyRed}{RGB}{180, 0, 0}
\definecolor{MyYellow}{RGB}{180, 180, 0}
\newcommand{\cmark}{{\textcolor{MyGreen}{\ding{51}}}}%
\newcommand{\xmark}{{\textcolor{MyRed}{\ding{55}}}}%
\title{Exploring easy boosts for lidar semantic scene completion}
\name{\begin{tabular}{c} 
      Tetiana Martyniuk$^{1,2}$, Jonathan Seele$^{1,3}$, Alexandre Boulch$^{1,2}$, \\ 
      Gilles Puy$^{1,2}$, Renaud Marlet$^{1,2,4}$, Raoul de Charette$^1$ 
      \end{tabular} }
\address{$^1$ Inria, France \quad $^2$ valeo.ai, France \\
         $^3$ ETH Zurich, Switzerland \quad 
$^4$ LIGM, CNRS, Univ Gustave Eiffel, ENPC, IP Paris, France}

\begin{document}
%
\maketitle
\begin{abstract}
This paper investigates ``free lunch'' strategies to boost the performance of lidar semantic scene completion (SSC) without requiring complex architectural redesigns. 
We first demonstrate that endowing input point clouds with semantic pseudo-labels from off-the-shelf segmentors significantly improves the performance of existing architectures. 
By evaluating these models against an oracle,
we establish that high-quality semantic priors are a  
primary driver of mIoU gains.
Furthermore, we equip the input lidar scan with visibility information that distinguishes between \emph{empty} and \emph{unknown} spaces, 
which provides a secondary performance boost across the tested architectures.
Using these simple enhancements, 
we observe that older models remain competitive with state-of-the-art systems, and can even outperform them. 
Our code is available at https://github.com/astra-vision/SSC-Priors.

\end{abstract}

\section{Introduction}
\label{sec:intro}

Semantic scene completion (SSC) for autonomous driving has seen a surge of interest in recent years~\cite{semantickitti,sscbench,occ3d}. 
SSC consists in estimating the semantic 
volume of a scene given the sensor output (e.g., camera, lidar, radar). 
It is most often framed as the estimation of 
a voxel grid where each voxel is either occupied or empty, and each occupied voxel is labeled with a semantic class.

This work focuses on \textit{SSC from lidar}, which estimates the semantic occupancy of a scene from a single lidar scan.
For simplicity, we refer to it as SSC. 
Compared to methods using images or depth map as input, a lidar provides accurate but sparse geometry and lacks dense colorimetric information which is a major cue for semantic estimation.
In recent years, several SSC methods have been proposed with various inputs and architectures, ranging from fast bird's-eye-view (BEV) processing~\cite{lmscnet} to complex denoising diffusion models~\cite{diffssc}.
\begin{figure}
    \centering
    \includegraphics[width=\linewidth]{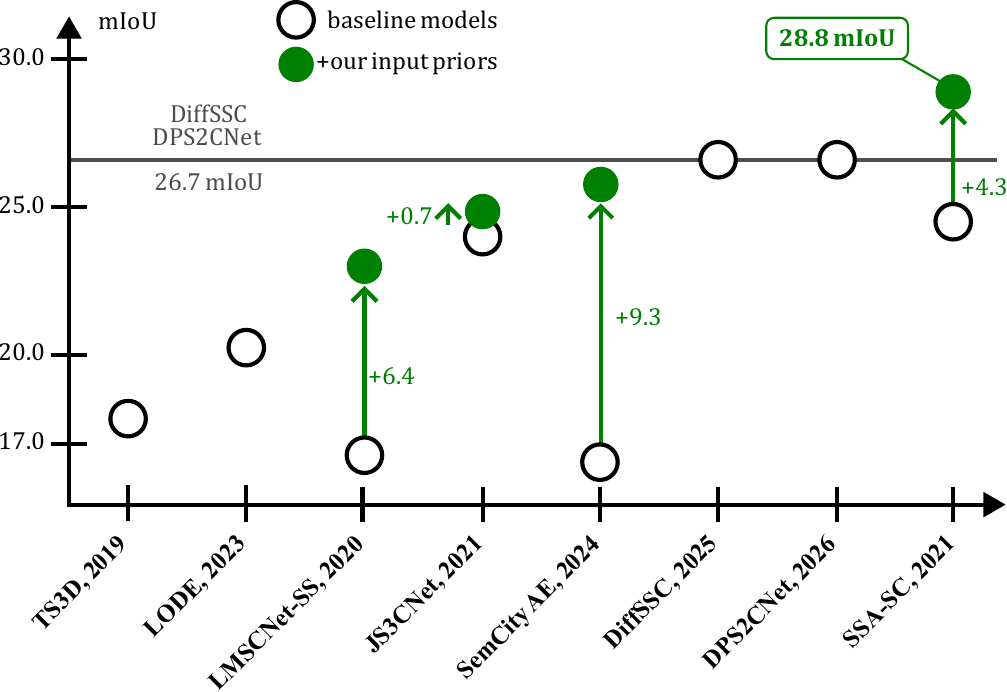}
    \caption{\textbf{Impact of input priors.} 
    We compare the baseline models against their versions endowed with our recipe. 
    It yields consistent gains across all backbones, propelling SSA-SC~\cite{ssasc} to 28.8 mIoU. 
    }
    \label{fig:teaser}
\end{figure}

Besides the voxelized lidar point cloud itself, some works 
benefit from additional priors derived from the input data, such as semantic or visibility information. 
These are typically learned concurrently~\cite{js3cnet,sscrs,ssasc}  
or derived from the sensor model~\cite{talos}. 
However, as the exact priors differ among methods and as their impact has not been systematically evaluated, it is unclear to what extent they contribute to the final SSC performance. 
Therefore, in this paper, we conduct a careful study leveraging either priors obtained from ground truth so as to establish theoretical upper bounds, or priors estimated from off-the-shelf techniques so as to demonstrate practical applicability. 
When equipping existing baselines with these estimated priors, we surprisingly find that older methods compete with the best current SSC techniques (see \cref{fig:teaser}).

Owing to the nature of the SSC task (jointly solving semantics and geometry), our study revolves around two specific priors: (a) \textit{semantics}, i.e., voxel semantic classes obtained from off-the-shelf point cloud segmentors~\cite{waffleiron,minkunet}, and (b)~\textit{visibility}, which provides the distinction between \emph{empty} and \emph{unknown} voxels by considering a simple ray casting derived from the sensor. 
Our study shows that using either of these priors boosts the performance of all tested SSC baselines. 
Furthermore, when combined, these priors are shown to be complementary, boosting the performance even further.

Our paper is an empirical study rather than a new method, so we depart from the classical "Method/Experiments" structure. 
\cref{sec:method} introduces and evaluates each prior individually, using both oracles and off-the-shelf estimators on two baselines (LMSCNet-SS~\cite{lmscnet} and SemCity-AE~\cite{semcity}). 
\cref{sec:experiments} then combines the priors and demonstrates their complementarity across four SSC architectures, including recent SOTA.

\noindent{}The lessons from our study are the following:
\setlist{nolistsep}
\begin{itemize}[noitemsep]
    \item Semantics is a bottleneck in current architectures: 
    using oracle input semantics establishes an upper bound for the SSC performance which improves upon baselines by a large margin;
    \item Off-the-shelf lidar semantic segmentation can provide useful semantic pseudo-labels and help mitigate the semantic gap with the oracle;
    \item Using sensor information, i.e., lidar lines of sight, to distinguish between \emph{empty} and \emph{unknown} voxels provides ``free'' boost to occupancy estimation.
\end{itemize}    
Applying these lessons jointly, we boost the performance of 4 models from the literature by +5.2 mIoU points on average, reaching the state of the art among reproducible methods.

\section{Related work}

\condensedparagraph{Lidar semantic scene completion.}
SSC was originally formulated as a densifying task from a depth map input~\cite{sscnet}. 
With the emergence of large-scale lidar datasets~\cite{semantickitti,sscbench,occ3d}, the existing TS3D~\cite{ts3d} method was adapted to lidar input by encoding point clouds as a voxelized TSDF, processed by a 3D CNN to output SSC. 
LMSCNet~\cite{lmscnet}, 
the first dedicated 
lidar SSC {method}, uses a multiscale loss with a 2D CNN instead, therefore turning the scene's height dimension into features. 
However, most of the follow-up SSC methods used discriminative 3D CNNs~\cite{s3cnet,sscrs,ssasc,js3cnet,completeandlabel}, sometimes combined with features from BEV representations~\cite{s3cnet,sscrs} or a semantic stream trained together~\cite{completeandlabel,js3cnet}. 
Instead of voxel representation, Local-DIF~\cite{localdif} directly encodes the point cloud with a point network~\cite{pointnet}, learning semantic completion via the ensembling of local implicit functions, at virtually infinite resolution. 
More recently, diffusion-based SSC were proposed~\cite{diffssc,ultralidar,semcity}, framing SSC as a conditioned-generation problem.
{These approaches builds on iterative generation processes for the distribution of the complete scene, represented as a dense set of points or a voxel grid.
They then use the lidar point cloud as guidance for the generation of the desired scene.}
Closer to our study, some methods derive additional semantic or geometric priors, plugged into the main SSC pipeline to boost performance. For example,~\cite{js3cnet,sscrs,ssasc} inject semantic priors {learned jointly with the SSC network}
while TALoS~\cite{talos} utilizes visibility ray tracing from {multiple time frame} lidar in a test-time adaptation setting. 
The benefit of these priors is, however, not fully evaluated. 
Instead, we propose a systematic study to quantify how semantic and visibility priors may boost existing SSC methods.

\condensedparagraph{Lidar semantic segmentation.}
{
3D semantic segmentation has made great progress over the recent years through various architecture types exploration~\cite{semsegreview}, from point-based~\cite{pointnet} and sparse convolution~\cite{minkunet} approaches, to methods specifically designed for automotive lidar using BEV projection~\cite{waffleiron}.
Recent advances on instance segmentation benchmark~\cite{semantickitti} have highlighted the benefit of lidar semantic segmentation for neighboring tasks.
We show in this study that it can also be used as an input prior beneficial for SSC.
}

\condensedparagraph{Using sensor information.}
{
Lidar are often processed as list of points, discarding the sensor information.
However, ray tracing from sensor to the point provides additional knowledge about the free regions of the scene, traversed by lidar rays, as well as about the unobserved regions behind occluded regions.
Such priors were shown to boost object detection~\cite{hu2020you}, lidar mapping~\cite{ding2019deepmapping} and even multi-frame test-time optimization for SSC~\cite{talos}. However, in the context of single-lidar-frame SSC its usage is not a common practice.
}

\section{Study of individual priors}
\label{sec:method}
\begin{figure*}[t]
    \centering
    \includegraphics[width=\linewidth]{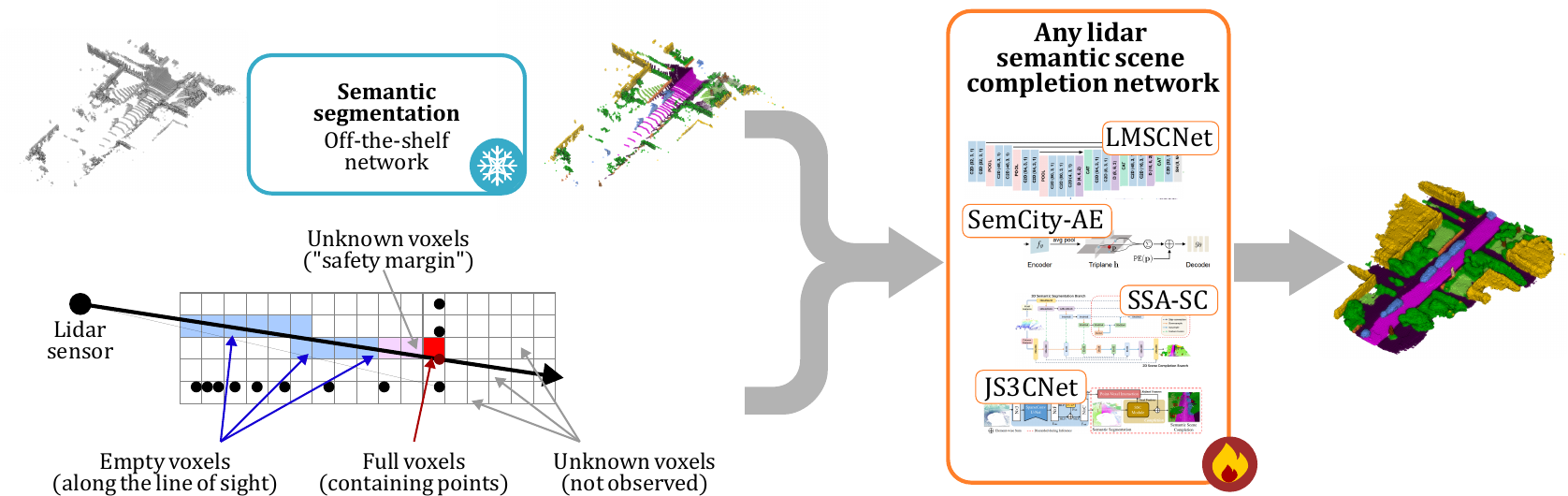}
    \caption{\textbf{Our recipe}. We study the impact of two input priors for lidar semantic scene completion: \textit{semantics} obtained from a frozen, off-the-shelf, point cloud segmentor, and \textit{visibility}, which provides the distinction between \emph{empty} and \emph{unknown} voxels by considering a simple ray casting derived from the sensor.
    }
    \label{fig:method}
\end{figure*}

In this section, we explore two recipes to boost the SSC performance, which can be applied readily to any existing
method.
The first one is the use of a semanticized input point cloud.
The second one is the use of visibility information to estimate empty 
and unknown  
voxels.

In order to show the wide applicability of these two recipes, we experiment on the SemanticKITTI benchmark with two networks: LMSCNet-SS~\cite{lmscnet} and the autoencoder of SemCity~\cite{semcity} (SemCity-AE), which we repurpose as a lidar SSC network. We experiment with two additional networks in \Cref{sec:experiments}. 

Formally, the task of lidar SSC is to start from a sparse lidar point cloud and output a completed volume where each occupied voxel is 
assigned one of the $C$ predefined semantic classes. 
Our method consists of providing \emph{input} 3D volumes to the SSC network augmented with the following information: for each voxel, we indicate if it is \emph{empty}, \emph{unknown}, or \emph{occupied} ( 
then it is assigned a corresponding semantic pseudo-label). 
The input is a 4D tensor constructed on a grid of voxels with one-hot 
vectors.
We denote this input 
by $\mathbf{V} \in \{0, 1\}^{X \times Y \times Z \times (C+2)}$, where $X$, $Y$, and $Z$ are the dimensions of the grid along the $x$, $y$, and $z$ axes.
Here, \emph{unknown} denotes voxels for which the lidar provides no reliable evidence — either unobserved (e.g., occluded) or within a safety margin around occupied voxels (see \cref{subsec:vis}).

\subsection{Semantic prior}

Several SSC methods leverage ground-truth (GT) semantic annotations of the input sparse point cloud either to construct auxiliary losses~\cite{ssasc} or to train an auxiliary network~\cite{diffssc}. 
On its own, semantic segmentation of sparse lidar point clouds is now a well-studied problem, with high-performing networks available off the shelf. 
In this section, we study the impact of augmenting $\mathbf{V}$ with semantic information derived from a semanticized lidar scan. 

Concretely, the tensor $\mathbf{V}$ is constructed as follows. First, all voxels in $\mathbf{V}$ are initialized with the \emph{unknown} label. 
Then, we identify all voxels containing at least one lidar point. 
For each of these voxels, its semantic label is determined by majority voting among points falling in this voxel.

\condensedparagraph{Oracle results.}
In order to estimate the maximum gain achievable with sparse semantic information at the input, we use the ground-truth annotated lidar point cloud to construct $\mathbf{V}$. The results in \Cref{tab:lidar_pseudo_labels} show that the semantic estimation of the \emph{completed} point cloud (mIoU metric) is significantly enhanced: +13.8 mIoU pts for LMSCNet-SS and +16.3 mIoU pts for SemCity-AE. 
Interestingly, the performance is above the recent method DiffSSC \cite{diffssc}, which reaches 26.7 mIoU. 
It highlights that LMSCNet-SS and SemCity-AE architectures struggle to estimate good semantics on their own, but this weakness can be greatly compensated with accurate sparse semantic segmentation at the input. 
We also remark a slight improvement in the completion task with a small boost of up to 0.7 pts in IoU.

\condensedparagraph{Pseudo-labels.}
We now investigate the performance achievable in a realistic situation where the 
input point cloud is semantically segmented with an off-the-shelf network. The results in \Cref{tab:lidar_pseudo_labels} show the performance achieved when using pseudo-labels obtained with a WaffleIron model~\cite{waffleiron}, with and without test time augmentation (TTA), or a MinkUnet model~\cite{minkunet} from the OpenPCSeg toolbox.

We first remark that the addition of pseudo-labels helps the semantic segmentation of the completed point clouds. 
Even though the gain is lower than with the oracle, it is still a significant improvement: up to +5.7 mIoU pts for LMSCNet-SS and +9.6 pts for SemCity-AE.
Second, we demonstrate that this performance boost is robust across different off-the-shelf segmentors. 
Both WaffleIron and MinkUNet provide significant gains over the baseline, confirming that the benefit of semantic priors holds true regardless of the specific state-of-the-art architecture used to generate them.

\begin{table}[t]
    \setlength{\tabcolsep}{2pt}
    \small
    \centering
    \begin{tabular}{c|clr|cc|cc}
    \toprule
    & \multicolumn{3}{c|}{Lidar labels} & IoU\% & $\Delta$ & mIoU\% & $\Delta$\\
    & & model & \% & \\
    \midrule
    \multicolumn{7}{l}{\emph{LMSCNet-SS}~\cite{lmscnet}}\\
    (a) & \xmark & none [retrained] & & 55.6 & -    & 16.6 & - \\

    (b) & \cmark & WaffleIron~\cite{waffleiron} & 68.0 & 55.6 & \textcolor{MyGreen}{+0.0} &  21.7 & \textcolor{MyGreen}{+5.1} \\
    (c) & \cmark & MinkUNet~\cite{minkunet} & 70.0 & 55.1 & \textcolor{MyRed}{-0.5} & 22.3 & \textcolor{MyGreen}{+5.7}\\
    (d) & \cmark & WaffleIron~\cite{waffleiron}+TTA & 70.3 & 55.6 & \textcolor{MyGreen}{+0.0} &  21.9 & \textcolor{MyGreen}{+5.3} \\
    \rowcolor{black!10}
    (e) & \cmark & GT        & 100.0 & 56.3  & \textcolor{MyGreen}{+0.7} & 30.4  & \textcolor{MyGreen}{+13.8}\\
    \midrule
    \multicolumn{7}{l}{\emph{SemCity-AE}~\cite{semcity}}\\
    (f) & \xmark & none & & 53.9 & -    & 16.4 & - \\

    (g) & \cmark & WaffleIron~\cite{waffleiron} & 68.0 & 54.4 & \textcolor{MyGreen}{+0.5} & 25.5 & \textcolor{MyGreen}{+9.1} \\
    (h) & \cmark & MinkUNet~\cite{minkunet} & 70.0 & 54.2 & \textcolor{MyGreen}{+0.3}& 25.5 & \textcolor{MyGreen}{+9.1}\\
    (i) & \cmark & WaffleIron~\cite{waffleiron}+TTA & 70.3 & 54.4 & \textcolor{MyGreen}{+0.5} & 26.0 & \textcolor{MyGreen}{+9.6} \\
    \rowcolor{black!10}
    (j) & \cmark & GT                         & 100.0 & 54.6 & \textcolor{MyGreen}{+0.7} & 32.7  & \textcolor{MyGreen}{+16.3}\\
    \bottomrule
    \end{tabular}
    \caption{\textbf{Semantic prior} effect on LMSCNet-SS and SemCity-AE networks. }
    \label{tab:lidar_pseudo_labels}
\end{table}

\condensedparagraph{Conclusion.}
These experiments indicate that the lightweight networks used in this section have difficulties to predict semantics on their own.
Providing this information as input to the network, thanks to pseudo-labels, leads to 
a significant boost in semantic performance, without requiring any architectural change (except the required adaptions to accept the $C+2$ channels of the input tensor).

\subsection{Visibility prior}
\label{subsec:vis}

We now study the impact of using visibility information to precompute voxels which should be empty after completion. 
Specifically, for every lidar point $p$, we sample new points $q_i$ every $\delta = 20$ cm on the ray between the lidar center $o$ and $p$. These auxiliary points $q_i$ are assigned the label \emph{empty}. 
A voxel is then labeled as \emph{empty} if and only if it contains at least one empty 
point~$q_i$ and lies outside a safety volume around the occupied voxels. 
The safety volume is used to minimize discretization artifacts caused by grazing rays slightly penetrating occupied voxels. 
The safety volume is composed of all occupied voxels as well as all voxels in their $5\times5\times5$ neighborhood.
Crucially, we assign the \emph{unknown} label to these ``safety margin'' voxels (even if traversed by a ray, see \cref{fig:method}).
The results in \Cref{tab:lidar_vis_ablation} show that using this visibility prior improves the completion IoU by up to 2.4 points. 
We also notice a mild positive effect on the semantic segmentation performance.

Finally, we also experiment with the oracle visibility prior derived from the GT. 
The results in \Cref{tab:lidar_vis_ablation} show that perfect knowledge of empty space not only solves the completion task (reaching near-perfect IoU) but also significantly aids semantic segmentation (+13.6 mIoU), proving that geometric ambiguity is a major noise factor for semantic prediction.

\begin{table}[t]
    \setlength{\tabcolsep}{3pt}
    \small
    \centering
    \resizebox{\linewidth}{!}{
    \begin{tabular}{cc|cl|cc|cc}
    \toprule
    & Model & \multicolumn{2}{c|}{Visibility}   & IoU\% & $\Delta$ & mIoU\% & $\Delta$\\
    \midrule
    (a)& & \xmark & [retrained] & 55.6  & -    & 16.6  & - \\
    (b)& & \cmark & our vis. prior & 56.9 & \textcolor{MyGreen}{+1.3} & 17.3 & \textcolor{MyGreen}{+0.7} \\
    (c)& \multirow{-3}{*}{LMSCNet-SS~\cite{lmscnet}} & \cmark & GT & 98.5 & \textcolor{MyGreen}{+42.9} & 30.2 & \textcolor{MyGreen}{+13.6} \\ 
    \midrule
    (d) & & \xmark & none    & 53.9	& - & 16.4 & - \\
    (e) & & \cmark & our vis. prior       & 56.3 & \textcolor{MyGreen}{+2.4} & 17.4 & \textcolor{MyGreen}{+1.0}\\
    (f) & \multirow{-3}{*}{SemCity-AE~\cite{semcity}}& \cmark &  GT        & 80.5 & \textcolor{MyGreen}{+26.6} & 28.7 & \textcolor{MyGreen}{+12.3}\\
    \bottomrule
    \end{tabular}
    }
    \caption{\textbf{Visibility prior} effect on LMSCNet-SS and SemCity-AE networks. }
    \label{tab:lidar_vis_ablation}
\end{table}

\subsection{Our proposed recipe}

We have seen in this section that augmenting the input tensor $\mathbf{V}$ with either semantic or visibility information significantly improves the performance of 
SSC networks.
We propose to combine both ingredients and study the benefit of synergic recipe in the next section.

\section{Combining the priors on SOTA methods}
\label{sec:experiments}

Having established in \cref{sec:method} that each prior individually boosts SSC performance, we now combine them and evaluate the joint recipe across four architectures: LMSCNet-SS~\cite{lmscnet}, SemCity-AE~\cite{semcity}, SSA-SC~\cite{ssasc}, and JS3C-Net~\cite{js3cnet}. 
The input tensor V is prepared with both semantic pseudo-labels and visibility information.
Note that $C=19$ for SemanticKITTI.

\begin{table}[t]
    \setlength{\tabcolsep}{2pt}
    \small
    \resizebox{\linewidth}{!}{%
    \begin{tabular}{@{}l|c@{}c@{}|c|cc|cc@{}}
    \toprule
    & \multicolumn{3}{c|}{\bf Ours} &  & & &\\
     & \multicolumn{3}{c|}{\bf Input priors} &  & & &\\
    \multicolumn{1}{c|}{\multirow{-2}{*}{Model}} & \multicolumn{2}{c|}{Sem.} & Vis. & \multicolumn{2}{c|}{\multirow{-2}{*}{IoU\% $\Delta$}} & \multicolumn{2}{c@{}}{\multirow{-2}{*}{mIoU\% $\Delta$}} \\
    \midrule
    SemCity-AE~\cite{semcity} &\xmark &  & \xmark  & 53.9  & -    & 16.4 & - \\
    LMSCNet-SS~\cite{lmscnet} &   \xmark & [retrained] & \xmark     & 55.6 & -    & 16.6 & - \\
    TS3D~\cite{ts3d} in~\cite{semantickitti} &  \xmark & [paper] & \xmark  & 50.6  & & 17.7 &  \\
    LODE~\cite{lode} & \xmark & [paper] & \xmark  & 51.2 &&20.2 & \\ 
    
    \rowcolor{blue!15}
    LMSCNet-SS~\cite{lmscnet} & \cmark & WI-TTA & \cmark        & 57.4 & \textcolor{MyGreen}{+1.8} &  23.0 & \textcolor{MyGreen}{+6.4} \\
    
    JS3C-Net~\cite{js3cnet} &          \xmark & [paper] & \xmark   & 57.0 & -    & 24.0 & - \\
    SSA-SC~\cite{ssasc}  &      \xmark & [paper] & \xmark       & 58.3 & - & 24.5 & - \\

    \rowcolor{blue!15}
    JS3C-Net~\cite{js3cnet} & \cmark & WI-TTA  & \cmark         & 56.7 & \textcolor{MyRed}{-0.3} & 24.7 & \textcolor{MyGreen}{+0.7} \\
    \rowcolor{blue!15}
    SemCity-AE~\cite{semcity} & \cmark & WI-TTA & \cmark & 56.7 & \textcolor{MyGreen}{+2.8} & 25.7 & \textcolor{MyGreen}{+9.3} \\
    DPS2CNet~\cite{liu5333789dual} & \xmark & [paper] & \xmark & \bf 60.8 & & 26.7 & \\
    DiffSSC~\cite{diffssc} &    \xmark & [paper] & \xmark                  & 60.3  & & 26.7 & \\ 
    \rowcolor{blue!15}
    SSA-SC~\cite{ssasc} & \cmark & WI-TTA       & \cmark   & 59.1 &\textcolor{MyGreen}{+0.8} & \bf 28.8 & \textcolor{MyGreen}{+4.3} \\

    \midrule
    
    \rowcolor{black!10}
    LMSCNet-SS~\cite{lmscnet} & \cmark & Oracle & \cmark    & 57.7 & \textcolor{MyGreen}{+2.1} & 31.0 & \textcolor{MyGreen}{+14.4} \\
    \rowcolor{black!10}
    SemCity-AE~\cite{semcity} & \cmark & Oracle & \cmark  & 56.4 & \textcolor{MyGreen}{+2.5} & 33.5 & \textcolor{MyGreen}{+17.1} \\
    \rowcolor{black!10}
    JS3C-Net~\cite{js3cnet} & \cmark & Oracle & \cmark            & 57.8 & \textcolor{MyGreen}{+0.8} & 33.7 & \textcolor{MyGreen}{+9.7} \\
    \rowcolor{black!10}
    SSA-SC~\cite{ssasc} & \cmark & Oracle & \cmark               & 59.8 & \textcolor{MyGreen}{+1.5} & 40.1 & \textcolor{MyGreen}{+15.6} \\

    \midrule
    \rowcolor{black!10}
    \multicolumn{8}{@{}l@{}}{\emph{Other methods from the literature}}\\
    \rowcolor{black!10}
    \multicolumn{4}{@{}l|}{SemCity refine {\scriptsize(GT occupancy leakage, see \cref{subsec:quanti})}} & \cancel{60.7} & & \cancel{25.6} & \\
    \rowcolor{black!10}
    \multicolumn{4}{@{}l|}{S3CNet~\cite{s3cnet} {\scriptsize(no code)}} & 57.2 &  & 33.1 & \\
    \rowcolor{black!10}
    \multicolumn{4}{@{}l|}{SCPNet~\cite{scpnet} {\scriptsize(non-reproducible. Cf. Sec.~5.2.1~\cite{liu5333789dual})}} & 50.2 & & 37.6 & \\
    \rowcolor{black!10}
    \multicolumn{4}{@{}l|}{TALoS~\cite{talos} {\scriptsize(multi-frame test-time optim.)}} & 56.1 & & 39.3 & \\
    \bottomrule
    \multicolumn{8}{@{}l}{\emph{WI: WaffleIron~\cite{waffleiron}; TTA: test-time aug. on lidar semantics;}}\\
    \multicolumn{8}{@{}l}{\emph{Oracle: ground-truth lidar semantic labels.}}
    \end{tabular}
    \begin{tikzpicture}[overlay, remember picture, shorten >=1pt, shorten <=1pt]

    \draw [o->, black] (0,1.6) to [bend right=-30] (0.1,-0.30);
    \draw [dashed, o->, black] (0,-0.30) to [bend right=-30] (0,-2.15);

    \end{tikzpicture}}
    
    \caption{\textbf{SemanticKITTI benchmark (validation set).} Green and red numbers refer to improvements over baseline scores thanks to addition of our input priors. }
    \label{tab:semkitti}
\end{table}

\begin{figure}[t]
\centering
\setlength{\tabcolsep}{1pt}
\small
\begin{tabular}{@{}cc@{}c@{}}
&(a) Input points & (b) Ground truth \\
&\includegraphics[width=0.48\linewidth, trim={2cm 4cm 2cm 3cm},clip]{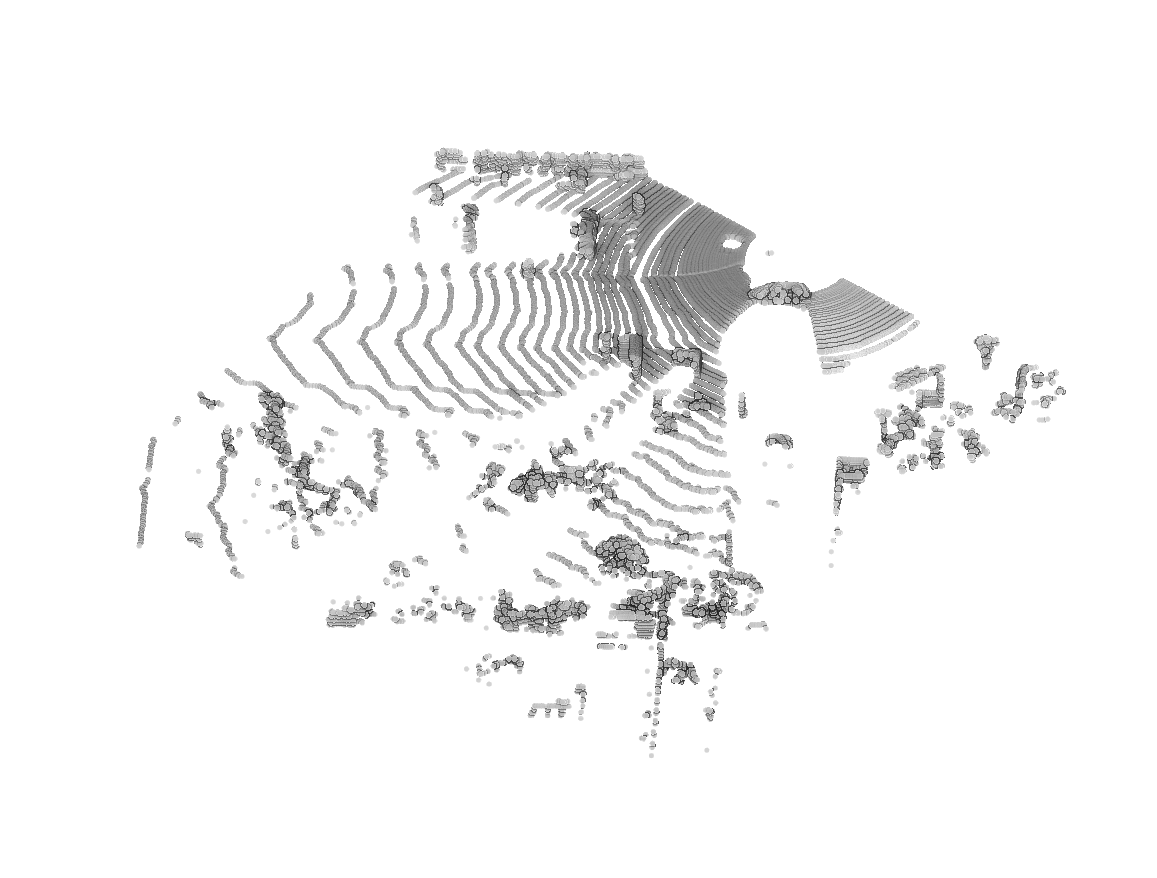} &
\includegraphics[width=0.48\linewidth, trim={2cm 4cm 2cm 3cm},clip]{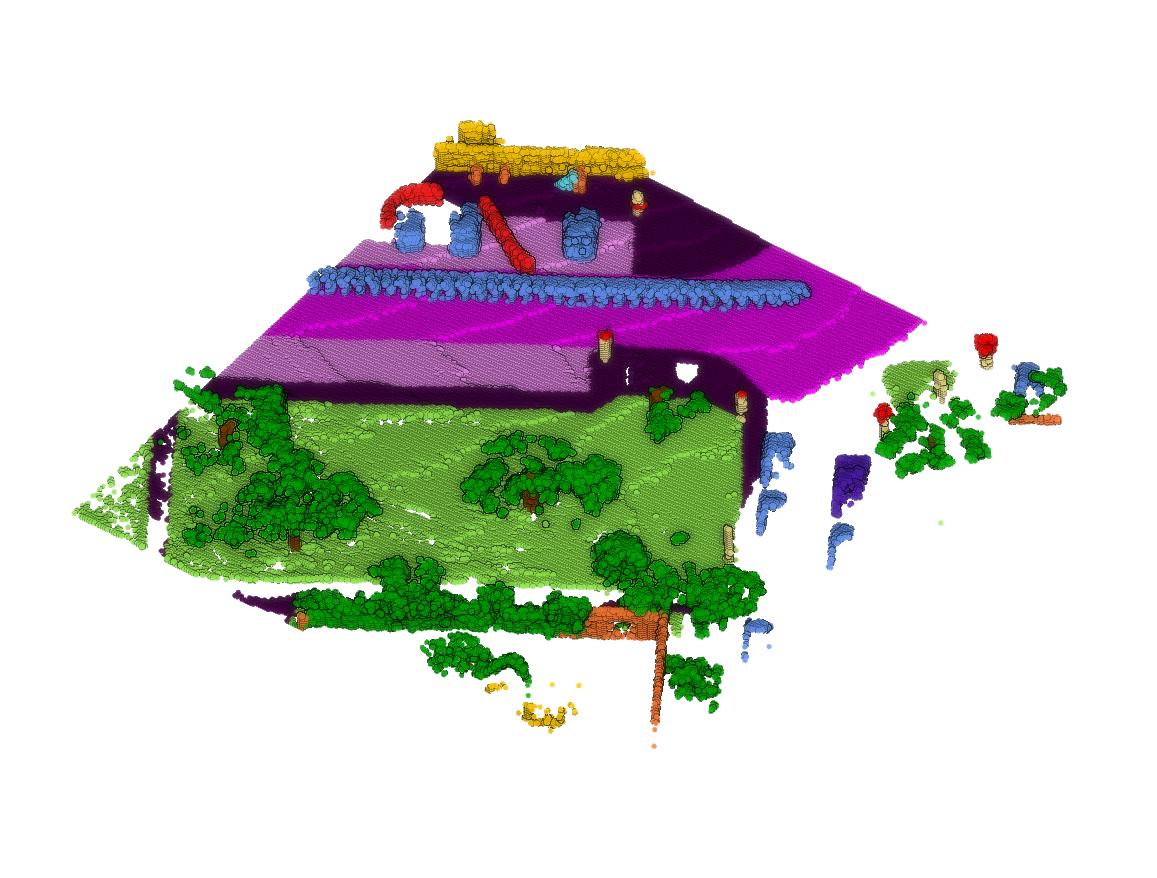} \\
\midrule

& (c) Original SSC & \makecell{(d) Original + ours \\(WI+TTA labels)} \\

\rotatebox{90}{\quad SemCity-AE } &
\includegraphics[width=0.48\linewidth, trim={2cm 4cm 2cm 3cm},clip]{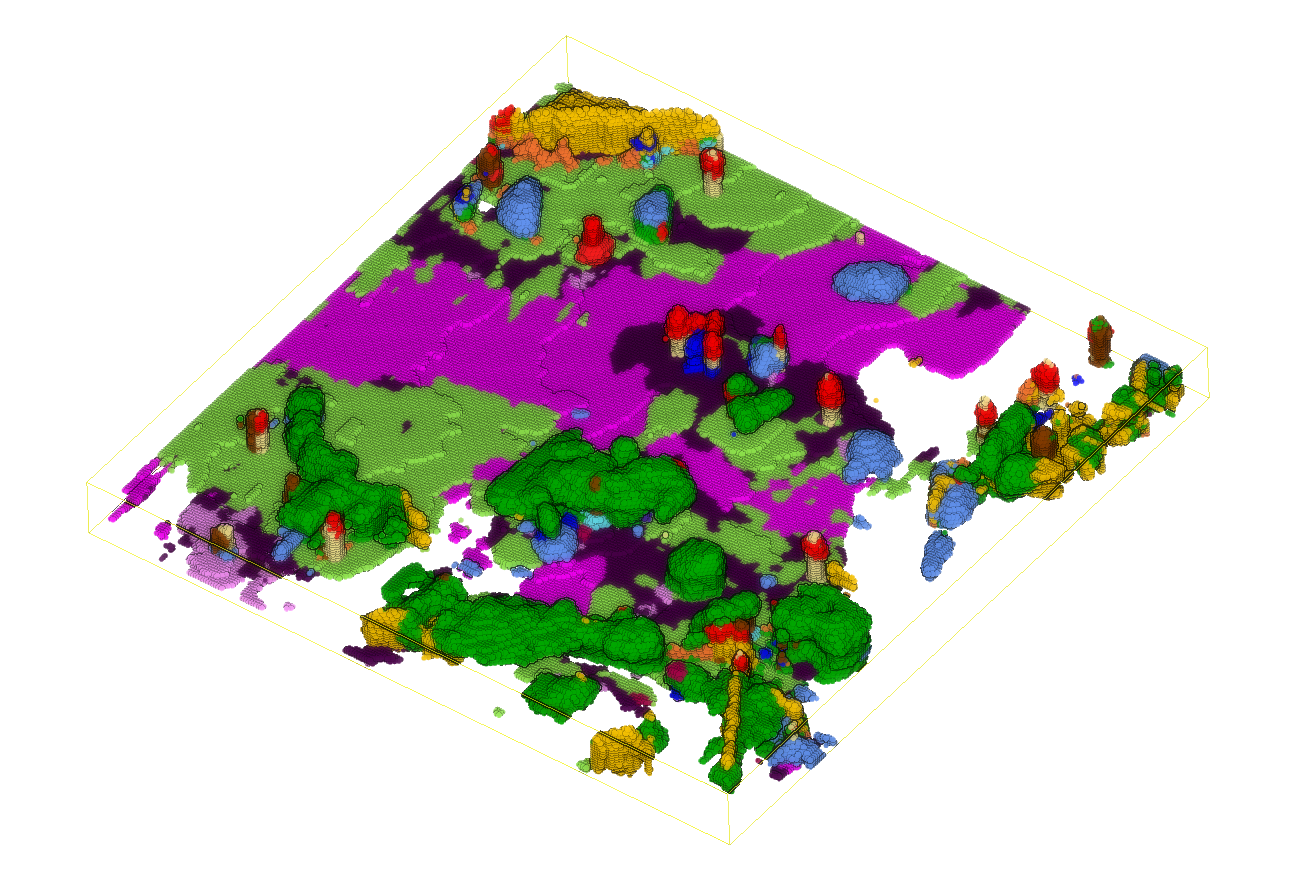} &
\includegraphics[width=0.48\linewidth, trim={2cm 4cm 2cm 3cm},clip]{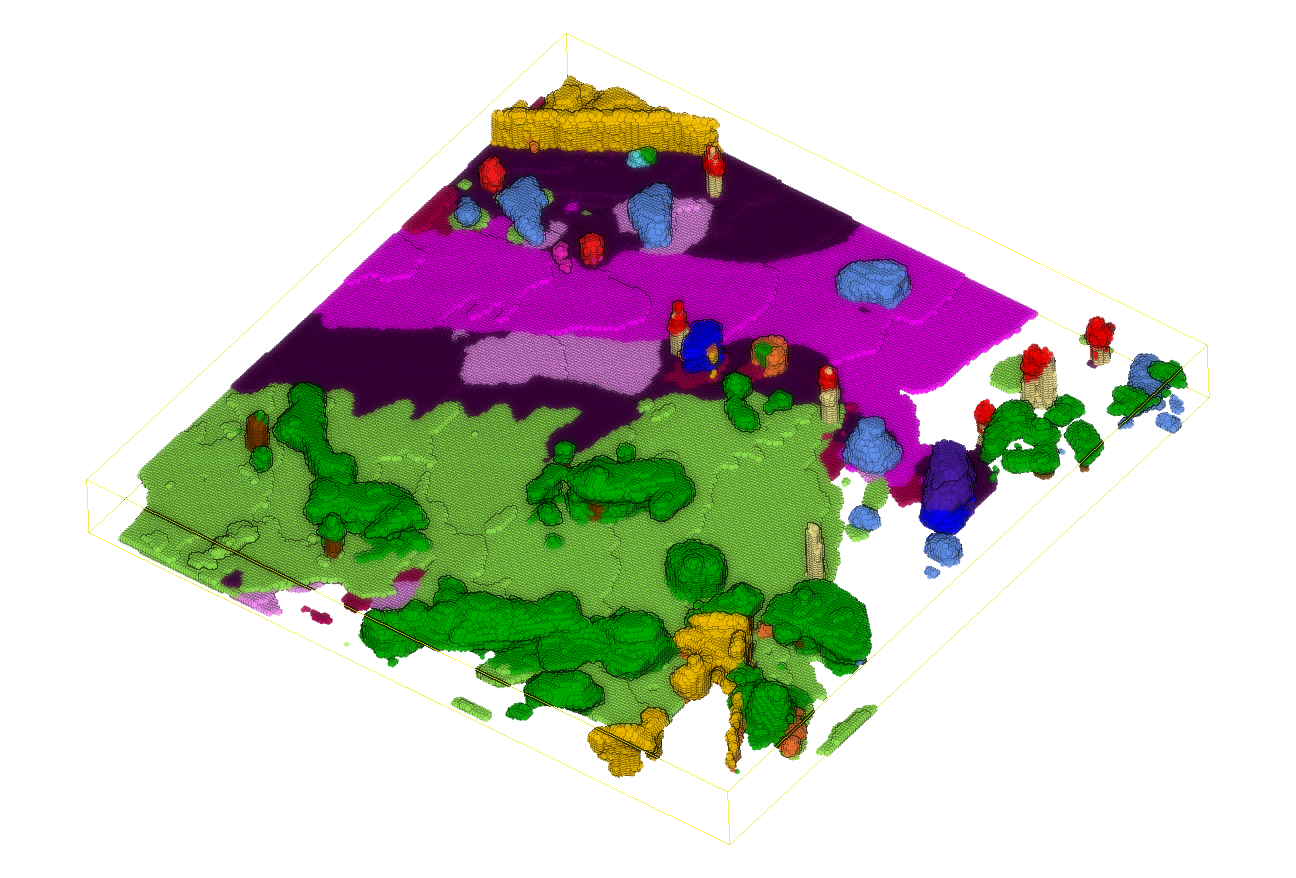} \\

\rotatebox{90}{~~LMSCNet-SS } &
\includegraphics[width=0.48\linewidth, trim={2cm 4cm 2cm 3cm},clip]{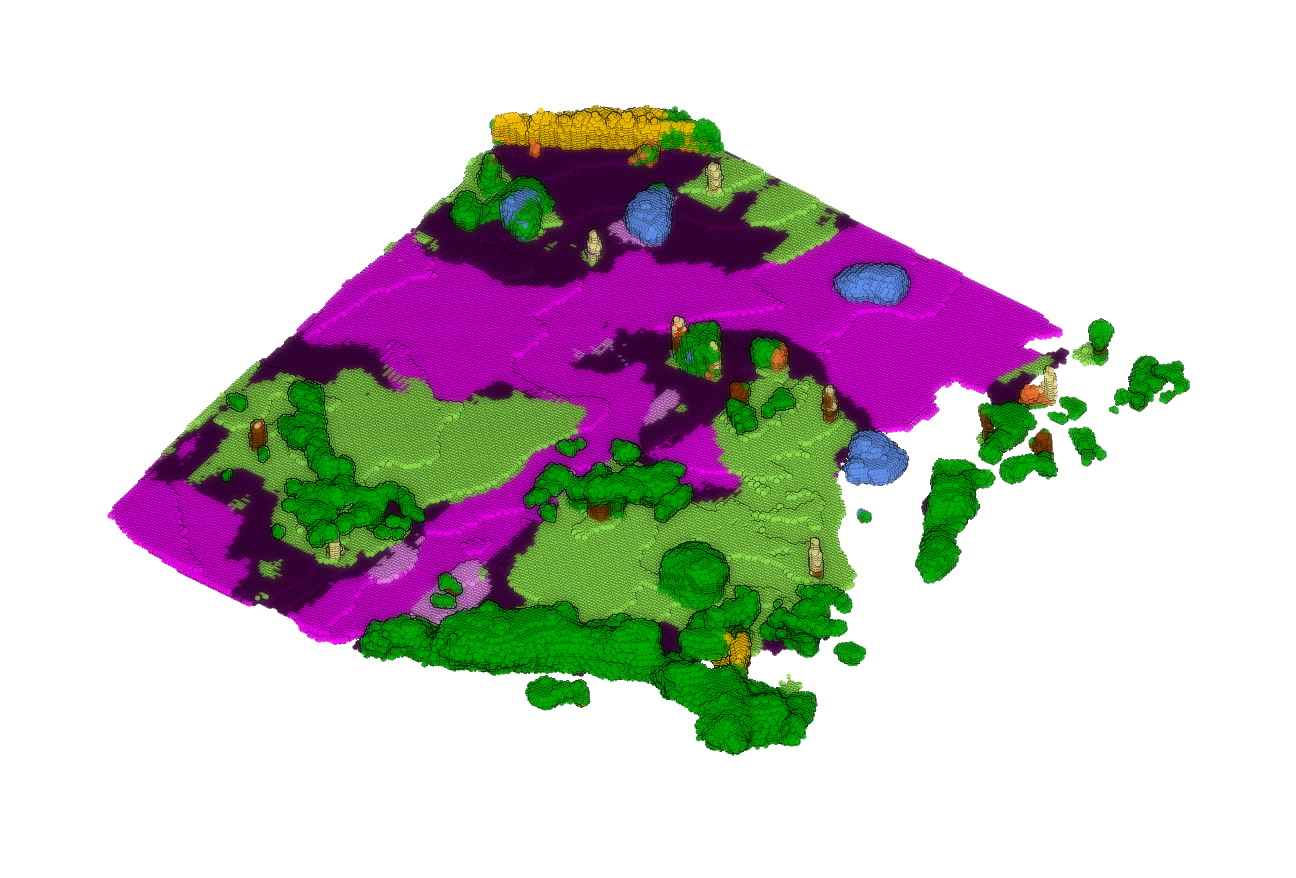} &
\includegraphics[width=0.48\linewidth, trim={2cm 4cm 2cm 3cm},clip]{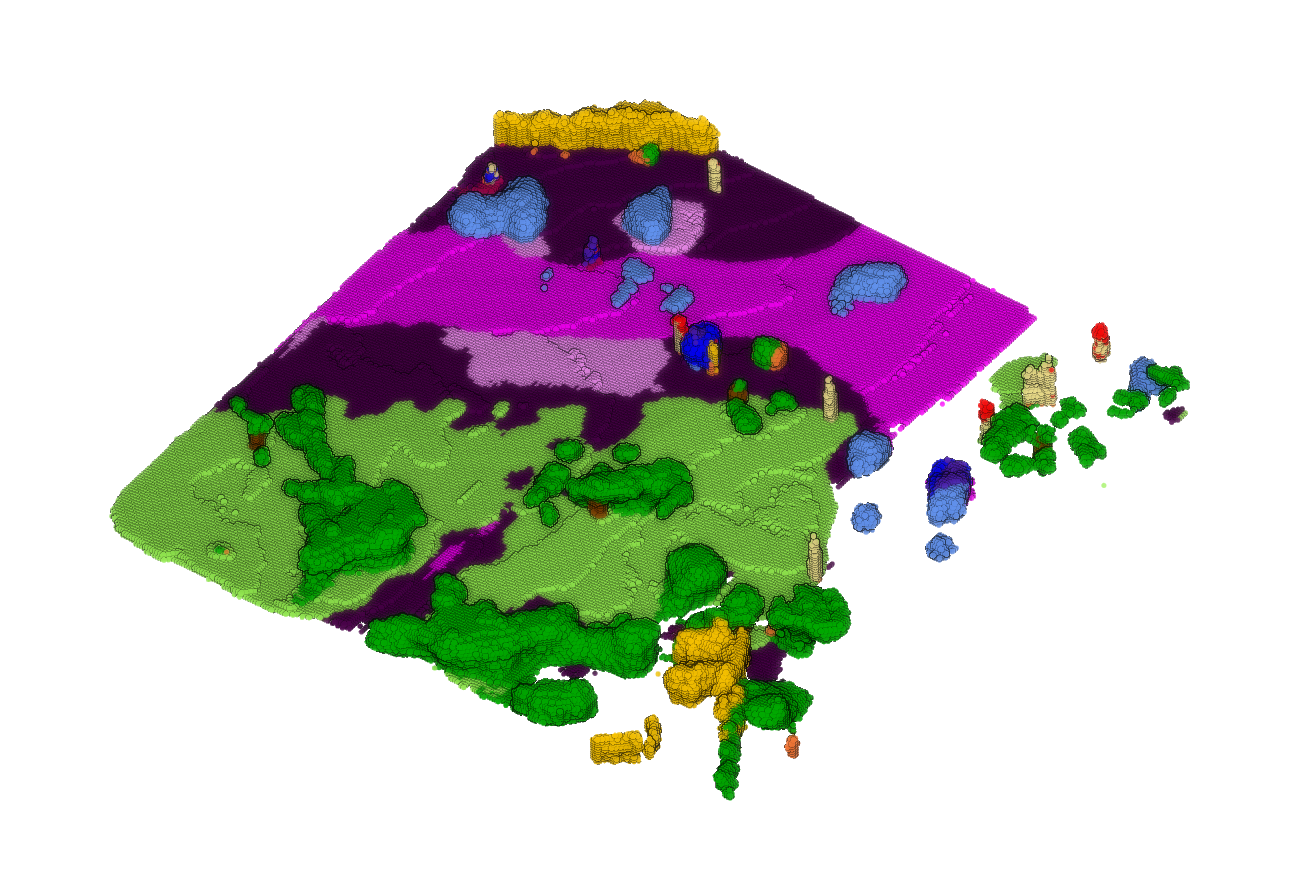}
\\

\rotatebox{90}{\quad\quad SSA-SC } &
\includegraphics[width=0.48\linewidth, trim={2cm 4cm 2cm 3cm},clip]{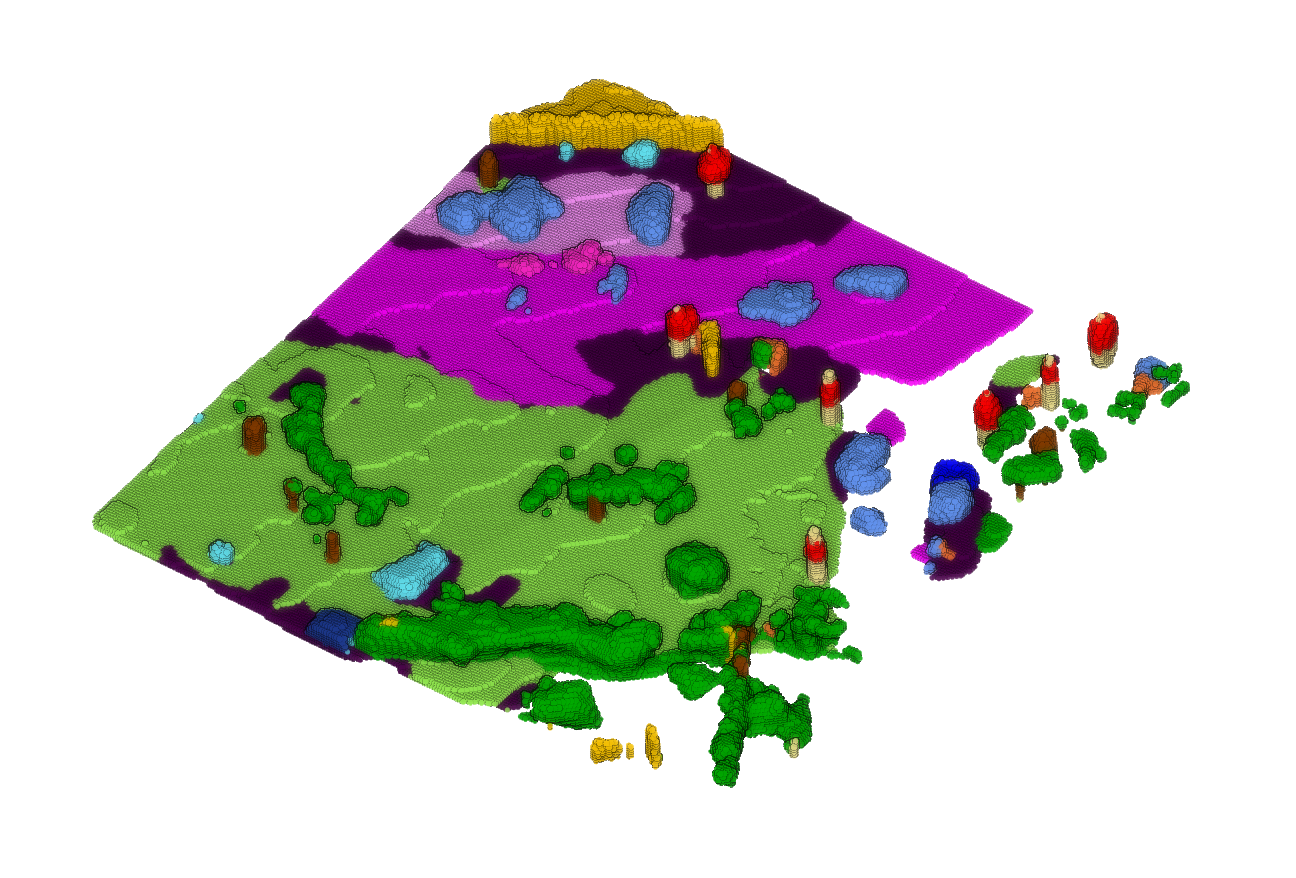} &
\includegraphics[width=0.48\linewidth, trim={2cm 4cm 2cm 3cm},clip]{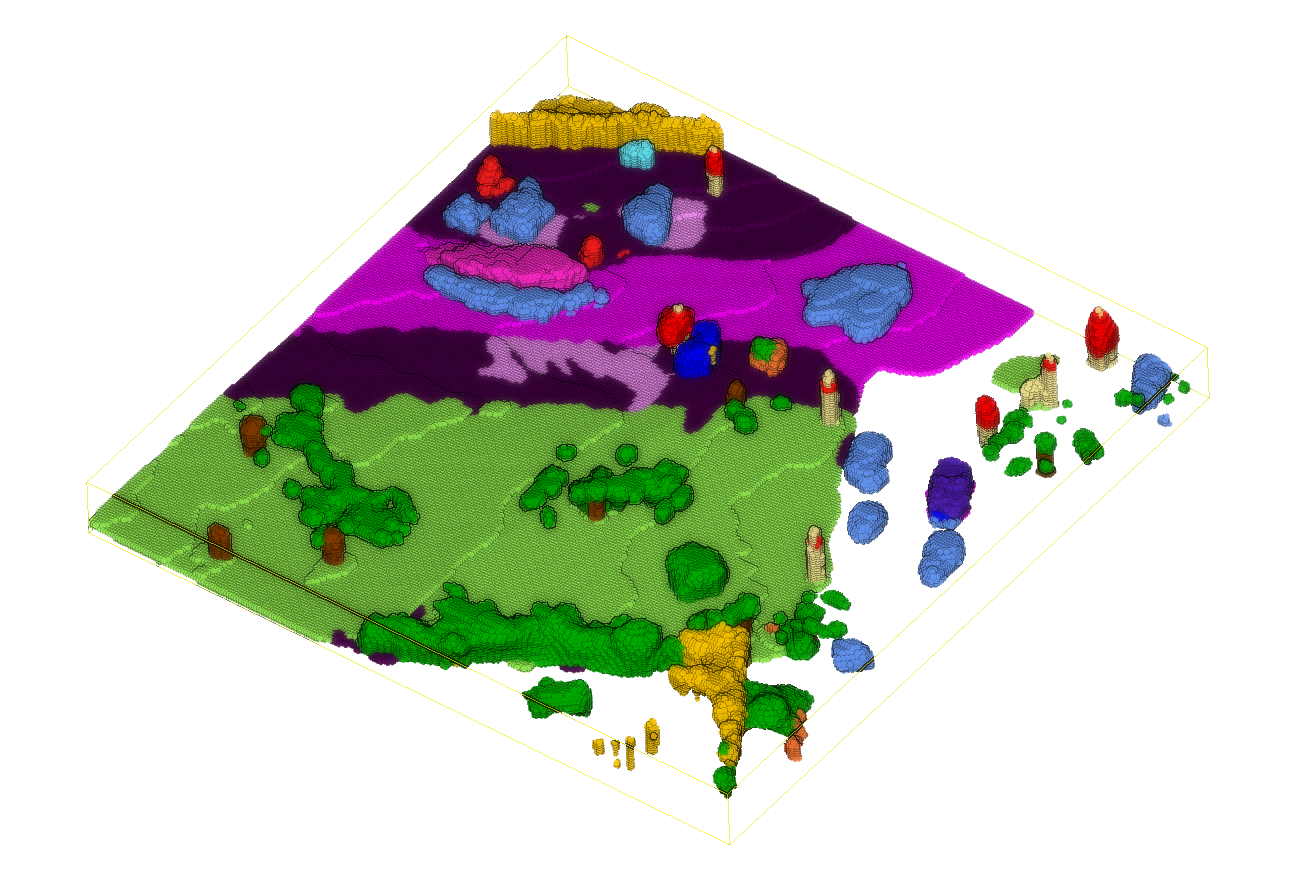} \\

\end{tabular}
\caption{
\textbf{Qualitative results on SemanticKITTI validation set.}
Unlike baselines (c), which operate on raw lidar input (a) and produce noisy, fragmented boundaries, our recipe (d) yields more structured and cleaner outputs for SemCity-AE, LMSCNet-SS, and SSA-SC, closer to the ground truth (b).
}
\label{fig:placeholder}
\end{figure}

\subsection{Implementation details}
\label{subsec:implementation}

\condensedparagraph{SemCity-AE.}
The original SemCity autoencoder was designed to encode dense ground-truth point clouds. 
We repurpose it for the SSC task by providing it with our augmented sparse tensor $\mathbf{V}$ as input and adapting the size of the embedding layer accordingly. 
The corresponding baseline, which has no access to semantic or visibility information, is obtained by providing voxel-wise 2-dimensional one-hot vectors as input, encoding only whether a voxel is occupied by at least one point from the raw lidar scan or not.

\condensedparagraph{LMSCNet-SS.}
LMSCNet-SS originally receives binary tensors of size $X \times Y \times Z$ and processes them as 2D tensors of spatial size $X \times Y$ with $Z = 32$ channels. In our case, the tensor $\mathbf{V}$ is treated as a 2D tensor of spatial size $X \times Y$ with $(C+2)\cdot Z$ channels. We just modify the first network layer to match the number of channels of the input tensors.

\condensedparagraph{SSA-SC.}
We change the input width of the BEV U-Net from $32+32$ input channels (occupancy + point feature channels) to $(C+2) \cdot 32 + 32$, as the 2D grid input to the network goes from binary occupancy to one-hot semantics with \emph{empty} and \emph{unknown} classes.
Both setups share the same PointNet and 3D segmentation branches,
while our ``enhanced input'' is only received by the 2D completion network.

\condensedparagraph{JS3C-Net.}
Its architecture requires two updates to leverage our semantic and visibility information. For the semantic segmentation module, we concatenate our one-hot encoded semantic pseudo-labels to the original per-point input features. 
For the SSC module, we enhance the input with our derived visibility information via an additional channel set to~1 when the corresponding voxel is empty, and 0 otherwise.

\subsection{Quantitative results}
\label{subsec:quanti}
\Cref{tab:semkitti} presents the quantitative evaluation on the Semantic-KITTI validation set~\cite{semantickitti}.
We compare the original performance of SemCity-AE, LMSCNet-SS, SSA-SC, and JS3C-Net to their performance when improved with our recipe. We leverage WI-TTA for the semantic pseudo-labels. We also experiment using ground-truth semantic labels on the input sparse lidar point clouds.

\condensedparagraph{Performance boost generalization.}
Our results demonstrate that endowing the models with explicit semantic and visibility priors yields consistent and significant improvements across all tested architectures: (a)~SemCity-AE, a lightweight autoencoder, improves the most, increasing the mIoU score by +9.3 pts and IoU score by +2.8 pts; (b)~LMSCNet-SS, a well-established baseline, improves by +6.4 mIoU pts, bringing this efficient architecture closer to today's SOTA; 
(c)~even SSA-SC and JS3C-Net, which 
already perform well in their vanilla settings, 
also benefit from our recipe (+4.3 and +0.7 mIoU points, respectively).

\condensedparagraph{SOTA reproducible performance.}
Most notably, applying our recipe to the SSA-SC network achieves 28.8 mIoU.
This establishes a new state of the art among fully reproducible, single-frame methods, outperforming recent 
approaches such as DiffSSC~\cite{diffssc} and DPS2CNet~\cite{liu5333789dual} (both at 26.7 mIoU).
While certain methods in the literature report higher scores (e.g., SCPNet at 37.6 mIoU or TALoS at 39.3 mIoU), these remain either closed-source (S3CNet), non-reproducible (SCPNet, cf.~Sec.~5.2.1 in~\cite{liu5333789dual}), erroneous (SemCity with SSC refinement)\footnote{
In the official code, the location of the non-annotated unobserved voxels in the ground truth 
are used during refinement, therefore leaking information about the location of the surfaces, the actual information to reconstruct.}, or rely on multi-frame test-time optimization (TALoS).
Our approach provides a plug-and-play alternative.

\condensedparagraph{The oracle headroom.}
Finally, our semantic oracle evaluations reveal that there is still a significant 
room for improvements with better semantic segmentation of the input sparse lidar point cloud.
For instance, SSA-SC reaches an impressive 40.1 mIoU under perfect semantic conditions.
The remaining gap between WI-TTA prediction and the oracle confirms our hypothesis: as off-the-shelf semantic segmentors continue to improve over time, the performance of these completion models --- even the lightweight models --- will continue to rise ``for free''.
Notably, adding our visibility prior on top of the GT semantic oracle still yields +0.8 mIoU on SemCity-AE (\cref{tab:lidar_pseudo_labels} row j vs. \cref{tab:semkitti} SemCity-AE Oracle), confirming the two priors remain complementary even when one is perfect.

\subsection{Qualitative results}
\Cref{fig:placeholder} shows the SSC outputs for Semcity-AE, LMSC\-Net-SS  and SSA-SC compared to their respective baselines.
Operating only on sparse lidar input
(\cref{fig:placeholder}a), the original baseline reconstructions (\cref{fig:placeholder}c) 
generate wrong semantics, and have noisy or fragmented class boundaries.
In contrast, when endowed with our input priors (\cref{fig:placeholder}d), the models produce cleaner and more structured outputs: continuous surfaces such as roads (purple) become fully connected, vegetation aligns better with the GT, and the poles with the traffic signs (right corner of the scene) are 
more reliably restored.
\section{Conclusion}
We presented a plug-and-play recipe that significantly boosts lidar SSC by enhancing input representations with off-the-shelf semantic pseudo-labels and visibility priors. 
Our extensive evaluation demonstrates that these simple input augmentations allow 
baselines to perform on par with more recent and complex state-of-the-art methods. 
This study highlights that semantic bottlenecks can benefit from the advances in point semantic segmentation, using off-the-shelf networks.

\condensedparagraph{Acknowledgments}
{\sloppy
This work was granted access to HPC resources of IDRIS under allocations AD011012883R4, AD011014484R2, AD011015497R1 by GENCI.
\par}


\bibliographystyle{IEEEbib}
\bibliography{strings,refs_short}

\end{document}